# Deciding About Semantic Complexity of Text By DAST Model

MohammadReza Besharati [a,1] and Mohammad Izadi [a,2]

[a] *Department of Computer Engineering, Sharif University of Technology, Tehran, Iran*

**Abstract.** Measuring text complexity is an essential task in several fields and applications (such as NLP, semantic web, smart education, etc.). The semantic layer of text is more tacit than its syntactic structure and, as a result, calculation of semantic complexity is more difficult than syntactic complexity. While there are famous and powerful academic and commercial syntactic complexity measures, the problem of measuring semantic complexity is still a challenging one. In this paper, we introduce the DAST model, which stands for "Deciding About Semantic Complexity of a Text". DAST proposes an intuitionistic approach to semantics that lets us have a well-defined model for the semantics of a text and its complexity: semantic is considered as a lattice of intuitions and, as a result, semantic complexity is defined as the result of a calculation on this lattice. A set theoretic formal definition of semantic complexity, as a 6-tuple formal system, is provided. By using this formal system, a method for measuring semantic complexity is presented. The evaluation of the proposed approach is done by a set of three human-judgment experiments. The results show that DAST model is capable of deciding about semantic complexity of text. Furthermore, the analysis of the results leads us to introduce a Markovian model for the process of common-sense, multiple-steps and semantic-complexity reasoning in people. The results of Experiments demonstrate that our method outperforms the random baseline with improvement in better precision and competes with other methods by less error percentage.

**Keywords.** Text, Semantic Complexity, intuitions, human-judgment experiments

## 1. Introduction

Complexity leads to overheads and on the contrary, text simplicity is usually desirable. The complexity of text is an important feature for numerous applications and systems. Most applications require text simplicity, such as: text readability assessment, some NLP problems (including text simplification, summarization, comprehension, interpretation, classification and clustering, data sets preprocessing, style detection, concept drift detection, topic modeling, automatic abstracting and etc.), foreign language education, smart education, semantic web problems and applications,

[1] Mohammad Reza Besharati, Department of Computer Engineering, Sharif University of Technology, Azadi Avenue, Tehran, Islamic Republic of Iran, E-mail: besharati@ce.sharif.edu, Corresponding Author. Tel: +98 21 6616 6699

[2] Dr. Mohammad Izadi, Department of Computer Engineering, Sharif University of Technology, Azadi Avenue, Tehran, Islamic Republic of Iran, E-mail: izadi@sharif.edu, Corresponding Author.

cognitive science and engineering, automatic editors, machine understanding of text, etc.

A text could be viewed as a composition of syntactic and semantic elements and characteristics. So it is reasonable to consider two principal types of text complexity: syntactic complexity and semantic complexity.

The problem of measuring Semantic complexity of a text is still a challenging one. Most of previous approaches use simple (word level, vocabulary lists, etc.), statistical (sentence length, number of propositions, etc.) or syntactical (surface features, sentence structure, etc.) measures for semantic complexity. These methods could not cover the all aspects of the problem and reach the ultimate depth of semantic features of a text.

In this paper, we introduce the DAST model which stands for "**D**eciding **A**bout **S**emantic Complexity of a **T**ext". In this model, an intuitionistic approach toward semantics lets us have a well-defined definition for semantic of a text and its complexity.

Concisely, we consider semantics and meaning as a construction, lattice[3] or system of realities (=intuitions). Any well-meaning text represents and references to a combination of entities, things, objects, events, affairs, facts, concepts, cognitions, affections or any other sort of basic realities and intuitions. So we could aggregately and abstractly consider semantic as a combination and construction of basic realities and intuitions (This manner of semantic definition is a constructive and intuitionistic one). Figure 1 illustrates the overall process of the proposed model and solution.

In the following section, we have a brief survey on relevant literature. Then, in the third section, the proposed model and solution is provided and an example of semantic complexity calculation is presented in fourth section based on that. In the fifth section, an evaluation of proposed solution is reported. The evaluation methods involved a set of three human judgement experiments. Discussion about proposed solution and experiment results are the next stage. After all, a conclusion ends this paper.

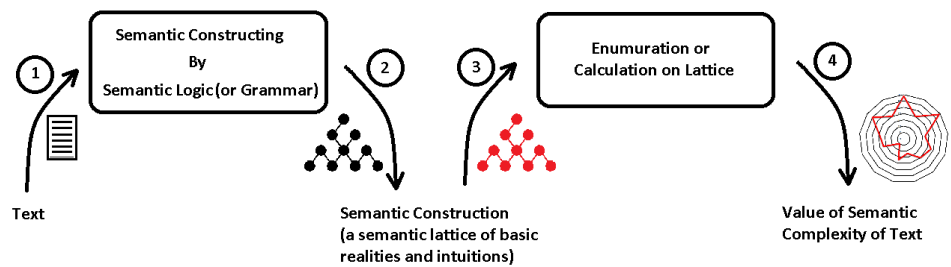

**Figure 1-** Overall process of the proposed model and solution

---

[3] Lattices are mathematical models for constructions in intuitionistic semantics.

## 2. Related Works

Most researchers in the fields of text complexity, text difficulty and readability have done their best for the problem of syntactic complexity and few of them have approached the problem of semantic complexity.

There are two principal families of approaches to semantic complexity of text based on objective and subjective points of view.

### 2.1. Objective Approaches (Text Oriented)

Some approaches to semantic complexity emphasize text (as the object under study in the process of reading), so they don't consider the effects of subjective issues on semantic and meaning. A brief overview of these approaches is provided in Figure 2.

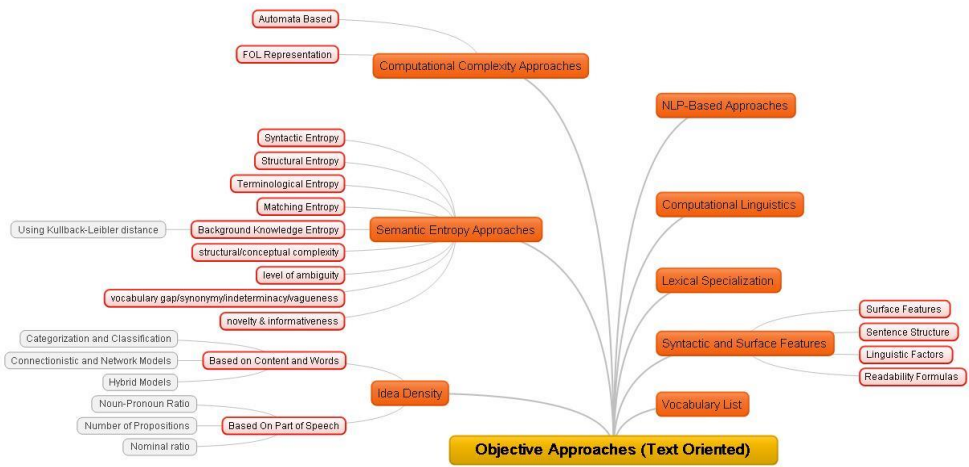

**Figure 2-** Objective Approaches (Text Oriented) to Semantic Complexity of Text

**Vocabulary list** or Vocabulary Load approach is based on a List of Complex\Simple Vocabulary terms (Badgett, 2010). Due to this approach, the more complex words are involved in the text, the more semantic complexity it has. A more complicated and robust manner based on this approach is to use a statistical language model as a measure of semantic complexity (Salesky & Shen, 2014): each word has a probability to occur in a text belonging to a particular class of difficulty (Collins-Thompson & Callan, 2004).

**Computational Linguistics** has an influential share in semantic complexity methods (Brunato, 2015), (De Clercq & Hoste, 2016), (Graesser et al., 2014), (Flor, Klebanov, & Sheehan, 2013), (Feng, Jansche, Huenerfauth, & Elhadad, 2010), and (Crossley, Skalicky, Dascalu, McNamara, & Kyle, 2017). In this approach, a set of linguistic features, factors and indices are incorporated and several semantic models are considered.

**Syntactic and surface features** are another means for semantic complexity calculation. This approach serves surface features (such as sentence length, letter count, syllable count, type-token ratio and word frequencies (Mühlenbock, 2013)), morphology and syntax factors (Dascalu, Dessus, Trausan-Matu, Bianco, & Nardy, 2013), readability formulas (Mühlenbock, 2013), sentence structure (Vajjala, 2015), and linguistic factors (Mühlenbock, 2013)) as mirrors of semantic complexity of a text. Some researchers try to find a readability assessment method solely based on syntactic, morphological, content-word and surface feature sets (for example, see (Dell'Orletta, Wieling, Cimino, Venturi, & Montemagni, 2014)).

Readability formulas have a long history of efforts that yield to a set of most numerical formulas. Constant values (based on experimental and corpus based evaluations) have a significant role in readability formulas. For example, here is The Flesch-Kincaid formula, which has been incorporated as a feature in some word processing software (Collins-Thompson, 2014):

$RG_{FK} =$
   $0.39 \cdot [(AverageWordsPerSentence)] + 11.8 \cdot [(AverageSyllablesPerWord)] - 15.59$

Type-Token Ratio (TTR) is an example of a surface feature. It is also known as vocabulary size divided by text length (V/N). This feature is a simple measure of lexical diversity (Kettunen, 2014). More lexical and vocabulary diversity may result in more semantic complexity (Hansson, Bååth, Löhndorf, Sahlén, & Sikström, 2016).

**Psycholinguistics Approaches** consider the findings of researches in language acquisition and psycholinguistics, cognitive linguistics and other related fields. These approaches capture cognitive aspects of word-meanings and create Psycholinguistics-based lexical features for assessment (Collins-Thompson, 2014). These types of lexical features include a word's average age-of-acquisition, concreteness, and degree of polysemy (Collins-Thompson, 2014).

**Idea density** is another approach which considers that contents containing more ideas to a given number of words are more complex (Mühlenbock, 2013). The complexity of relations between ideas is also another source of complexity. Some approaches consider number of entities and concepts behalf of ideas.

Ideas are measured in two different ways: based on content and words and based on part of speech. Both of the ways have their subdivisions which are depicted in Figure 2.

Another tool for idea density measurement is calculating average number of senses per word which is reported as a successful feature for readability assessment (Pilán, Volodina, & Johansson, 2014).

By Categorization and Classification of words (such as Content Word Classification (Mühlenbock, 2013)), Connectionist and Network Models (such as Semantic Cognition Model (Rogers & McClelland, 2004), WordNet Based Approaches

(Štajner, Evans, Orăsan, & Mitkov, 2012), by means of lexical chains and active chains (Brunato, 2015)), Hybrid Models (such as Saldo lexical-semantic network (Mühlenbock, 2013)), by measuring the longest path that a semantic network of a sentence contains (Vor Der Brück & Hartrumpf, 2007), Noun-Pronoun Ratio (Mühlenbock, 2013), Number of Propositions (Mühlenbock, 2013) (Vajjala, 2015) (Moraes, McCoy, & Carberry, 2014) (Brunato, 2015) and Nominal ratio (Mühlenbock, 2013), the semantic complexity of a text could be estimated.

In **Semantic Entropy Approaches**, an information theoretic point of view leads the semantic complexity definition (Freitas, Sales, Handschuh, & Curry, 2015). There are some entropy measures which let us calculate this type of semantic complexity (Freitas, 2015): Syntactic Entropy, Structural Entropy, Terminological Entropy, Matching Entropy, Background Knowledge Entropy (using Kullback-Leibler distance for text categorization), structural/conceptual complexity, novelty & informativeness, level of ambiguity and vocabulary gap/synonymy/indeterminacy/vagueness.

**Lexical specialization** is a mean for abstracting a set of special identities, ideas and meanings. Usually a specialized term has a higher semantic complexity than a common semantic one (Haase, 2016). So counting the special words and the common semantic words in a text can be a method for calculating some measures of semantic complexity.

**Computational Complexity Approaches** use a computational settlement for proposing a definition of semantic complexity. Some linguistics and cognitive science researchers argue that the brain capability of information processing and efficient computation has a significant role in human natural languages (Yang, Crain, Berwick, Chomsky, & Bolhuis, 2017)(Szymanik, 2016)(Gennari & Poeppel, 2003). So the underlying computational complexity of a sentence or word, which means the minimal computational device recognizing it (Szymanik & Thorne, 2017) or the computational complexity of satisfiability problem of its First-Order logic (FOL) representation (Thorne, 2012)(Pratt-Hartmann, 2004)(Third, 2006), could be considered as a measure for semantic complexity.

This literature contains some interesting questions (as research agenda questions), for example: What are the semantic bounds of natural languages or, in other words, what is the conceptual expressiveness of natural language? (Szymanik & Thorne, 2017).

**NLP-based approaches** use some techniques and analysis methods of NLP in order to measure semantic complexity. Distributional semantics and especially Latent semantic analysis (LSA) have a profound application in these approaches (Dascalu et al., 2013), (Hansson et al., 2016). LSA also has its application in modern readability measures (Landauer, 2011), (Graesser et al., 2014) and semantic complexity (Chersoni, Blache, & Lenci, 2016).

A thread and trend of research in machine learning based NLP is devoted to deciding about difficulty and complexity of text based on various syntactic, statistical and word-list features (for example, see (Kauchak, Mouradi, Pentoney, & Leroy, 2014), (Chauhan, Kaluskar, & Kulkarni, 2015) (Tseng, Chen, Chang, & Sung, 2019)).

These approaches try to construct a statistical model for text classification in order to predict text complexity with a higher accuracy in comparison with traditional readability formulas (Mühlenbock, 2013). Deep Learning is also used to train neural text difficulty classifiers (Filighera, Steuer, & Rensing, 2019). Some hybrid approaches use few features related to the semantics such as average number of sense per sentences (for example, see (Pilán, Vajjala, & Volodina, 2016) and (Pilán, 2018)).

Text Simplification methods (Coster & Kauchak, 2011)(Kauchak et al., 2014), (Paetzold & Specia, 2016)(Vajjala & Lucic, 2018) (especially features used in machine learning based simplification) are related to the problem of deciding about text complexity. Essay Evaluation (Zupanc & Bosnić, 2015) is another field which is technically related to the evaluation of text complexity. Some researchers also consider full-fledged machine-learning methods to approach the problem of text readability assessment (Vajjala & Lucic, 2018), (Naderi, Mohtaj, Karan, & Moller, 2019).

## 2.2. Subjective Approaches (Reader Oriented)

Subjective approaches rely on the fact that the semantic complexity has a subject side (the reader as a human). So we couldn't build up a proper semantic complexity metric or suite without considering the subject side. There are many approaches trying to model subject, his\her attitudes, capabilities and preferences. Each Subject has his\her cognitive, affective and real world which could affect the semantics and semantic complexity of text.

Some approaches bring attention to human interest and interestingness of stimulus materials (Mühlenbock, 2013) based on the relation between syntactic and surface features of text to human interest and interestingness. Also here are other approaches that use some content- and meaning-related features for interestingness. For example, the information in a text which is novel and complex yet, but still comprehensible could result in a peak of interest (Van Der Sluis, Van Den Broek, Glassey, Van Dijk, & De Jong, 2014). Level of proficiency, prior knowledge of materials addressed in the text and style and genre of the text (Mühlenbock, 2013) (Munir-McHill, 2013) are another metrics of subjective calculation of semantic complexity. The subject, itself, has a conceptual complexity of his\her mind (Mufwene, Coupé, & Pellegrino, 2017) which affects the process of semantic and meaning interpretation.

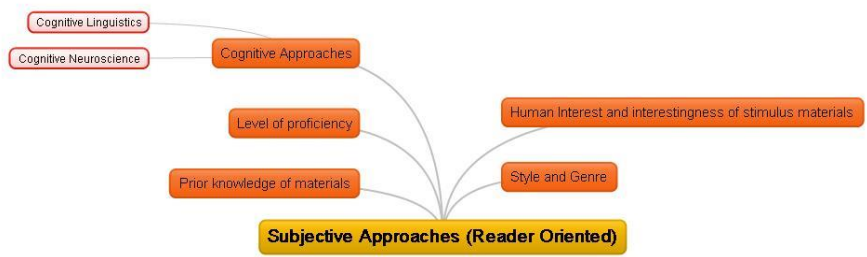

**Figure 3-** Subjective Approaches (Reader Oriented) to Semantic Complexity of Text

**Psychometrics** could be served as a measure for complexity. Because psychologic aspects of person and text could affect the semantics, Psychometrics could be considered as an indicator of subjective semantic complexity. For example, Integrative Complexity (IC) is considered as a measure of existence for multiple perspectives and connection of them in a reading setting (Robertson, Aiello, & Quercia, 2019).

Some **cognitive neuroscience** studies use the methods of this field to study semantic complexity. For example, Magnetoencephalography which is a functional neuroimaging technique is used as a means for semantic complexity measurements and experiments (Brennan & Pylkkänen, 2010). In such experiments the activity of reader's brain is measured and considered as an indicator of semantic complexity. In another research, it is found that the harder meaning tasks have caused higher transcranial magnetic stimulation (TMS) values in the brain (Ralph, Jefferies, Patterson, & Rogers, 2016).

Some experimental **cognitive linguistics** studies pay attention to semantic complexity calculations. For example, some sort of such studies design reading experiments and consider reading time as a measure of semantic complexity (Gennari & Poeppel, 2002).

## 3. The Proposed Model

The principal component of our proposed model is its manner of semantics. So, we first introduce our proposed notion of Semantic Logic. Based on it, we are able to construct lattices of intuitions.

### 3.1- Semantic Logic

Semantic Logic is an intuitionistic logic. Technically, it could be viewed as an axiomatic system on symbols with Brouwer–Heyting–Kolmogorov Interpretation for semantics (Sato, 1997). For our intention in DAST, it is sufficient to define Semantic Logic as a system consisting of 1) a set of symbols (behalf of basic intuitions), and 2) a set of rules on them. Each rule describes a symbol generation action: when the left-side symbolic structure of the rule is ready in the working memory, the right-side symbolic structure is generated and pushed to the lattice of symbols in the working memory (See Figure 4).

The manner and model of computation that is associated with Semantic Logic is similar to rewriting logics, generative grammars (especially when grammars be used for constructing a parse tree) and the LISP programming language. (For example, a working memory is an important element in these computing models and all of these computing models are symbol manipulating ones.)

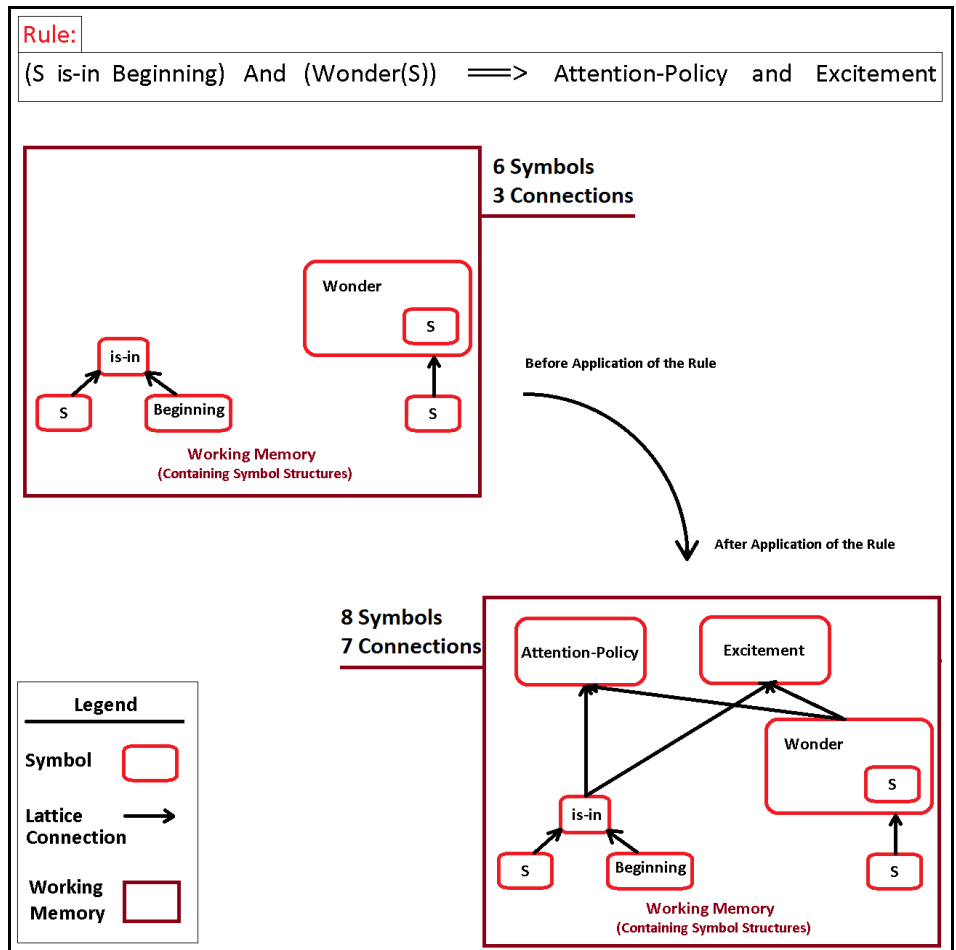

**Figure 4-** An example of a Semantic Logic rule and the effect of its application on working memory.

### 3.2- Semantic and Meaning Definition

In DAST, we consider text semantic and meaning as a closure, lattice or system of realities. Any well-meaning text represents and references to a combination of entities, things, events, affairs, facts, physics, concepts or any other sort of realities. So we could aggregately and abstractly consider semantic as a combination of realities.

Our realistic approach is not despite of conceptual one. We consider concepts and conceptual things as an important sort of realities. Albeit and Also, we have another types of realities, too.

In intuitionistic logic, our path towards realities is through intuitions. We build a construction of intuitions and intuitive values on behalf of the reality and to represent and reference to them. So, we consider semantic and meaning as a closure, lattice or system of intuitions and intuitive values.

This intuitionistic, realistic, conceptual and constructive approach lets us to have a formally defined model of semantic and meaning. Because of the computational nature of constructive and lattice-based approaches, we could easily define semantic complexity on this model of semantic.

**3.3- Semantic Complexity Definition**

Many things on a lattice could be counted and computed: height or depth, paths length, degrees, number of nodes, frequency of different involved constructive operators and etc.

By assigning and designating weights to elements of a lattice, we could compute overall or local values on it.

For a given text, based on any different set of principal intuitions, we may construct a different lattice as a semantic and meaning for that text. So we have two arguments for semantic calculation: given text and given set of principal intuitions.

For a given semantic of a text (a semantic lattice), based on any different configuration of weights and values on the lattice, we may compute a different complexity value. Also, the computation method and algorithm should be determined and be given before calculating complexity. So we have three arguments for semantic complexity calculation: given semantic lattice, given value configuration on it and given algorithm for complexity calculation.

Altogether, our model of semantic complexity could be defined as a 6-tuple formal system: (Text, Set of principal intuitions, semantic calculation algorithm, Semantic Lattice, lattice-value configuration schema, complexity calculation algorithm)

**3.4- Formal Definition of DAST Semantic Complexity Model**

So as to be able to compute the semantic complexity by this model, we need further refinements to achieve and define more concrete levels of this model. In this step, Formal definition for each component of this model is provided.

**(1) Text:** In computer science and engineering, strings are the formal representation of textual data and information. Some other generalizations such as regular expressions and grammars could be utilized for further refinements. Because of simplicity, it's sufficient for us to consider strings as a proper model of textual data. Rules, grammars and other higher level text-related structures could be modeled by other components in this model.

**(2) Set of principal intuitions:** A formal language (defined on an alphabet or set) is our formal representation of this component. Albeit, any other formal representation or formalism that could be able to model a set of mentioned things could be utilized. For the sake of simplicity, this formal language mechanism is selected.

**(3) Semantic calculation algorithm:** A Semantic Logic (a set of constructive-derivation rules on principal intuitions) would be provided for realization and behalf of this component. It could be viewed like a grammar on a symbolic alphabet. Logic systems and algorithms are equivalent mathematical objects.

**(4) The semantic lattice:** the resulting lattice that is constructed for the text by using semantic calculation algorithm.

**(5) lattice-value configuration schema:** This component could be realized and defined in rules and axioms of Semantic Logic (For example, by tagging the rules).

**(6) Complexity calculation algorithm:** Using lattice-values, complexity of text could be calculated.

**Definition 1.** A semantic complexity (CM) is a 6-tuple
CM = ⟨T, P, SA, L, V, CA⟩, where:

1. T is a string and representing the text-under-study.
2. P is a finite set of symbols on behalf of principal intuitions. All symbols in T are members of P. P might also have other members.
3. SA is an algorithm for constructing a lattice of symbols on behalf of semantic construction of the text. This algorithm consists of a set of symbol generation rules (SR) and a construction procedure (CP).
    a. SR ⊆ { b → c | b ∈ $2^{P*}$ ∧ c ∈ $2^{P*}$ }
    b. CP=
        i. For each r ∈ SR, if its left-side symbolic-structure is ready in working memory, generate its right-side symbolic-structure and connect it to the lattice of symbolic values.
        ii. Return to i.
4. L is a lattice of symbolic values which is constructed by SA algorithm.
    L is a set of pairs (i ,r), where:
    a. i ∈ ℕ ∧ r ∈ SR
    b. ∀ (i,r) ∈ L . (i>1) ⇒
        ∃ $(j,q_1),…, (j,q_n)$ ∈ L .
        (j=i-1) ∧ firstComponent(r) ⊆ ∪secondComponent($q_i$)
    5. V is an algorithm (= function) for computing a numeral value for each node of L based of its predecessors. So any sub-lattice in L is assigned to a numeral value. The type of V is:   ℙ(ℕ × SR) → ℕ
6. CA is an algorithm (= function) for computing a numeral value for each pair (L,V). The type of CA is:   ℙ(ℕ ×SR) × ℙ (ℙ (ℕ×SR) × ℕ) → ℕ

### 3.5- An application of semantic complexity

Abduction is a principal type of reasoning (Douven, 2011) (Y. Wang, 2007). Some cognitive and semantics researchers thought that its role in human thinking and human cognition is important.  An apparent difference between conventional computing (sometimes called Computing by Machines) and natural human cognition (sometimes called Computing by humans or brains) has root in this type of reasoning

(Y. Wang, 2007): we, humans, are able to do abduction, but the ability of conventional machines are limited to induction and deduction (two other types of reasoning).

Some new computational paradigms (such as semantic computing, cognitive computing, computing with words, perception based computing, perceptual computing, brain computing, etc.) have recently addressed and accepted the importance of abduction (Gust, Krumnack, Schwering, & Kühnberger, 2009) (P. Wang, 2019) . These paradigms are trying to exploit abduction in order to achieve higher level of understanding, intelligence and smartness in artificial and machinery computing. Some researchers are seeking about a non- von Neumann computing architecture without its limits and have found their missing and desired model of computation in human brains which serves abduction in its natural reasoning mechanisms (Gust et al., 2009) (Y. Wang, 2009).

Some researchers believe our human mechanism of abduction is working based on a principle of economy (Wirth, 1998). Abduction needs and works by hypothesizes and in practice of our daily abduction reasoning, we first select and accept simpler, more efficient and more natural explanations and hypothesizes.

So simpler hypothesizes are intended in abduction. Measuring Semantic complexity lets us to select the simpler choices and to build a mechanism of abduction by constructing semantic lattices for choice space members (or answer set members) and select the simpler one.

In some approaches for analogical reasoning (which incorporates in itself some sort of abductive reasoning (Tour, Boy, & Peltier, 2014)), the analogical generalizations with least structural complexity are preferred in a process of analogy generating task (Schmidt, Krumnack, Gust, & Kühnberger, 2014) (Besold, Kühnberger, & Plaza, 2017). In these approaches, least complexity is an approximation for good mappings in general analogical reasoning (Schmidt et al., 2014). Because the Constituent elements of an analogical mapping structure are concepts (Veale & Li, 2014), their approach to structural complexity is somehow similar with our semantic complexity notions (for example, see (Schmidt et al., 2014)). So the semantic complexity could have some potential applications in the field of analogical reasoning (for example, comparable with notion of "the complexity of the underlying mappings" in (Abdel-Fattah & Krumnack, 2013)).

## 4. An Example of Semantic Complexity Calculation

A syntactic simple but semantically somehow complex sentence is considered in this example. We call it as the sentence under study (SUS).

**SUS-1:**   See More, Feel More.

Some Semantic Logic axioms related to the semantic of this sentence is provided in the Semantic Logic-1 (see remaining). With these axioms (rules), we are able to construct the semantic lattice of SUS-1. In this example, Semantic complexity of SUS-

1 is addressed from cognitive meanings point-of-view. So, most of intuitions in the provided Semantic Logic is cognition and psychologic-related ones. (Anybody could construct him\herself Semantic Logic from his\her point of view. It is because this fact that the nature of semantic is a multi-view, multi-model, multi-modal one).

**Semantic Logic-1:**
1. (A => B) => (↑Quality(A) => ↑Quality(B))
2. See => [(Verb)]
3. Feel => [(Verb)]
4. [(Verb)] More => ↑Quality([(Verb)])
5. See => To-See
6. To-See => To-See-Drama | To-See-Documentaries
7. To-See-Documentaries => Wonder-Displaying | Sightlines | Knowing | To-Learn-Skills | To-Experience | To-Influence | Entertainment
8. To-See-Drama => Imagination | Fantasy | To-Learn-Skills | To-Experience | To-Influence | Storytelling | Promotion | Entertainment
9. (A => B) => (↑Quality(A) => ↑Quality(B))
10. Feel => To-Feel
11. To-Feel => Cognition | Affection
12. Cognition => Knowing | Intelligence | Expertise
13. Affection => To-Influence | Entertainment
14. [(Verb)] => ?[(Verb)]

By application of the axioms of provided Semantic Logic, the semantic lattice of SUS-1 is constructed (figure 5 to figure 7). Each edge of lattice is the result of application of one axiom. The number of applied axiom is written on each edge.

After the construction of semantic lattice, as mentioned in figure 1, we need a calculation of complexity values on lattice nodes in order to compute semantic complexity values. This calculation needs a schema. This schema guides us to configure nodes with values and calculate these values. In this example, we choose this simple value configuration and calculation schema on lattice nodes:

$$\begin{cases} \text{Leaf Complexity} = 1 \\ \text{Node Complexity} = 1 + \prod_{i=1}^{Number\ of\ Child-Nodes} Complexity(ChildNode_i) \end{cases}$$

The result of application of this schema is presented in figure 5 to figure 7. On each node, a value of its complexity is written. All these values in all levels of semantic lattice of SUS-1 could be attributed to SUS-1 as one of its semantic complexity indicator or measure. Albeit, overall complex measures and metrics also could be defined and calculated. Finally, The Most Complex Semantic Values of SUS-1 and their calculated complexity weights are summed up in figure 8.

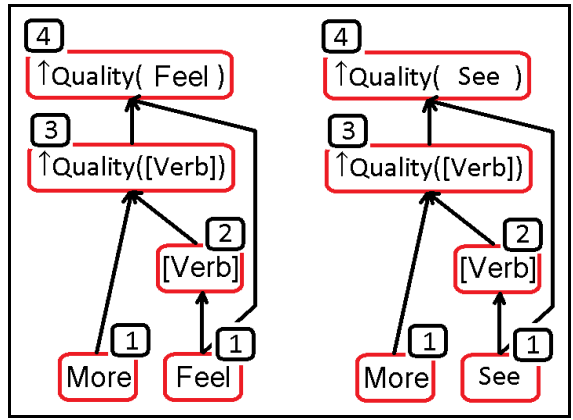

**Figure5 -** Part1 of the semantic lattice of SUS-1 and the configuration of complexity values on nodes.

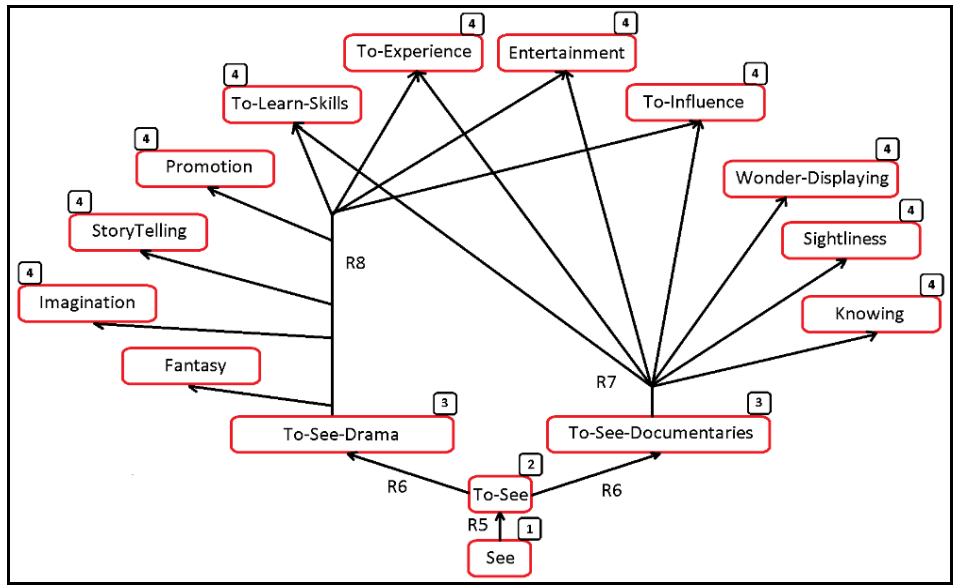

**Figure6 -** Part2 of the semantic lattice of SUS-1 and the configuration of complexity values on nodes.

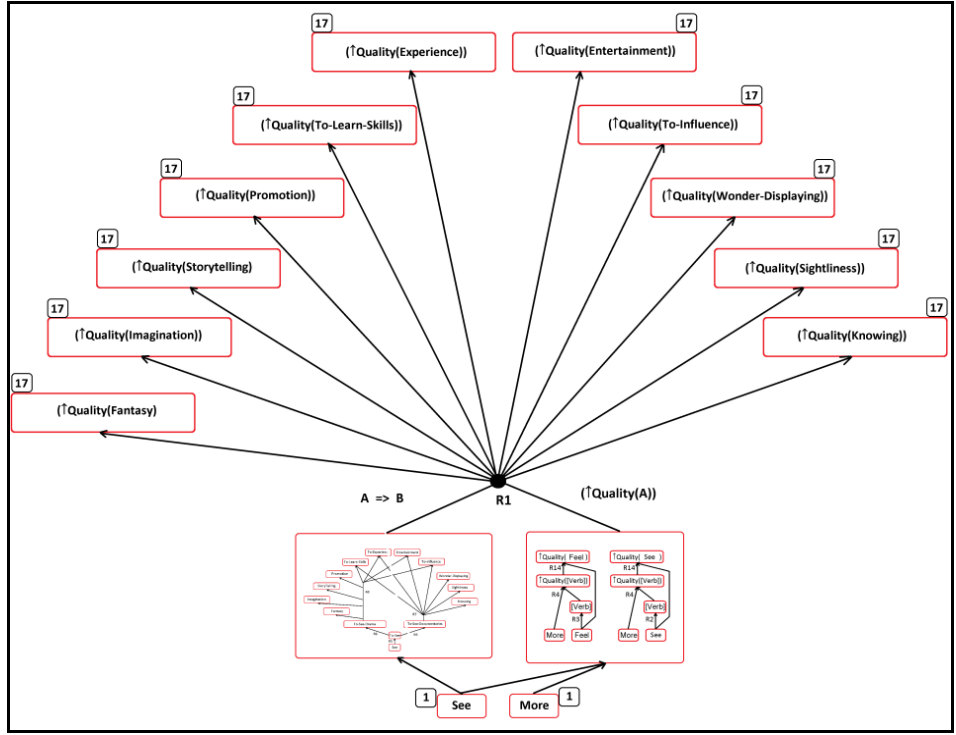

**Figure 7-** Part3 of the semantic lattice of SUS-1 and the configuration of complexity values on nodes.

These values in figure 8 indicate a notion of semantic complexity: The more value, the more complication and the more content. It's because of information combination and blending during multi-input derivations.

Complex semantic elements (in this notion, the elements with greater calculated values) could be interpreted as more informant semantic elements. It's because of combination of information in the structure of lattice during exposing and constructing each semantic element. Albeit, a logarithmic normalization could be done to avoid the impact of simple (with no information combination) reasoning. For example, by this calculation schema, a chain of N derivations results in an N+1 complexity value. A linear chain doesn't provide significant information combinations. So Log (N+1) is more proper than N+1. (Because Log (N+1) << N+1)

Each derivation could be considered as an epistemic event. So statistical, probabilistic or any other type of event analysis could be applied on semantic lattice. So besides structural complexity measures, Dynamical complexity measures (such as the sort of dynamic systems analysis or Kolmogorov-Sinai Entropy) could be considered and calculated.

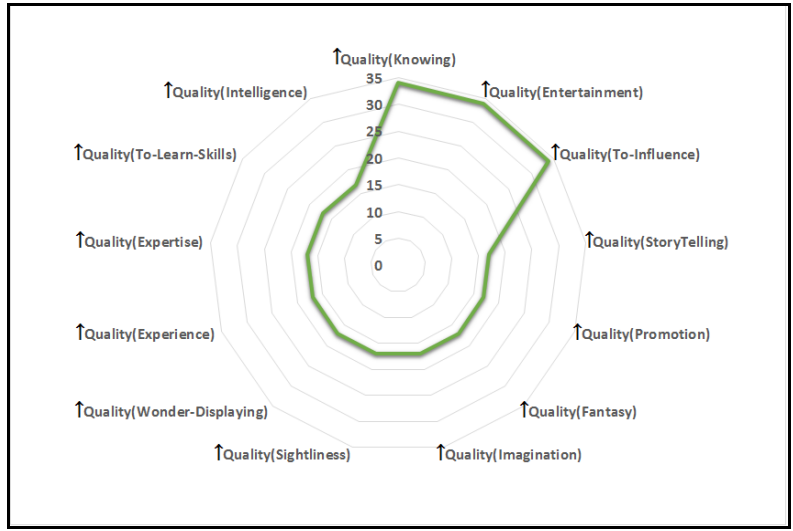

**Figure 8**- Most Complex Semantic Items and their Complexity Values for SUS1 and the resulting Semantic Space.

State transition Machine (or DFA) is also an accepted model of an important type of dynamic systems (particularly, Automation systems). So, derivation events could shape state-transition automation in order to provide a secondary calculation of complexity. Each state of this automation system could be interpreted as a class or stage of semantic complexity.

## 5. Modeling the notion of Overall Semantic Complexity of a Sentence

An analogy with a linear algebraic space with n-dimensions could contribute to a better elaboration of DAST notion of semantic complexity for a sentence. In this analogy, each semantic item of a sentence (for example: "increase in quality of entertainment", as a semantic item calculated from SUS1) could be viewed as a semantic dimension and Complexity value of that item (in this example: 34) could be viewed as a determination of semantic value on that semantic dimension. So as an overall result, the distance between the origin point and "semantic point" of a sentence in this semantic space could be considered as an overall complexity value of that sentence. A sentence with a calculated semantic point which is far from the origin point, is a semantic complex one.

A Human has an overall estimation for semantic complexity of a sentence. This overall estimation, judgment or understanding could be computationally modeled with above notion of overall semantic complexity of a sentence. So a CM (see Definition 1) constructs a semantic space, n semantic dimensions and calculates semantic items and complexity values for each of them.

**Definition 2.** Semantic space is an n-dimensional linear algebraic space which its dimensions are in behalf of semantic items defined by a CM.

**Definition 3.** Semantic point is a point in a semantic space which defined by determination of values on semantic dimensions regarding CM-calculated complexity values of semantic items of a text.

**Definition 4.** Overall semantic complexity of a sentence is the distance of its CM-calculated semantic point from the origin in the regarding semantic space.

In Definition 1, we defined CM by a set-theoretic definition. Here, we defined semantic space, semantic point and overall semantic complexity of a sentence. These definitions are linear algebraic notions. So a set-theoretic meta-model ( = definitions of CM) results in a new layer of definitions with a linear-algebraic meta-model of notions ( = Semantic Space, Semantic Point and overall semantic complexity). This weaving and relating of notions and models from different principal mathematical meta-models shapes a hybrid theoretic framework. This constitution is necessary for a capable and enough-complex computational model of real-world and humanistic notion of semantic complexity.

**Definition 5.** A semantic theory is made up of a domain of realities (hence basic intuitions) and some symbolic rules on them.

For the sake of easiness, we may estimate the overall semantic complexity of a text by calculating the number of its involving semantic theories. This first level estimation allow us a more simple and pragmatic evaluation of semantic complexity.

**Definition 6.** For a text, the number of involving semantic theories is a first level estimation of its semantic complexity:

DAST Semantic Complexity Index (DASTEX) = number of involving semantic theories

So, After all, we could see that these definitions provide a theory (based on intuitionistic philosophy and logic), a model (based on semantic lattices), a method (based on a basic algorithmic calculation) and a simple estimation technique (based on the concept of semantic theory). These elements and components shape a semantic complexity Framework, Called DAST.

## 6. Experiment and Evaluation

The DAST solution and model for semantic complexity of text is evaluated with three known methods of evaluation: 1) detailed example and case study, 2) Human Judge evaluation, and 3) Corpus Based Evaluation. The first is a well-known and general evaluation method for software engineering solutions (Runeson & Höst, 2009). The second is a classic evaluation method for NLP systems and solutions (Clark, Fox, & Lappin, 2010). The last one is a generally accepted method for NLP systems (Clark et al., 2010), computational linguistics (Ide, 2017) and Automatic Readability Assessment (ARA) solutions (Vajjala & Lucic, 2018).

**6.1- Detailed Example and Case Study**

We have seen a previous example in section 4. A more complex sentence under study (SUS) is chosen for this detailed example. The chosen SUS is a real one, in the sense that it is the beginning sentence of a TED talk about black holes[4].

**SUS-2:** How do you observe something you can't see?

With an intuitionistic approach to understanding the meaning and semantic of this sentence, 11 semantic theories (domains of realities and rules on them) are detected. Each semantic theory shares basic intuitions and rules with some other theories. The theories for SUS-2 and the dependencies between them are illustrated in figure 9. Under each theory, we have some Semantic Logic rules (like grammar rules on symbols of basic intuitions). Semantic Logic-2 is Semantic Logic of SUS-2.

**a)**

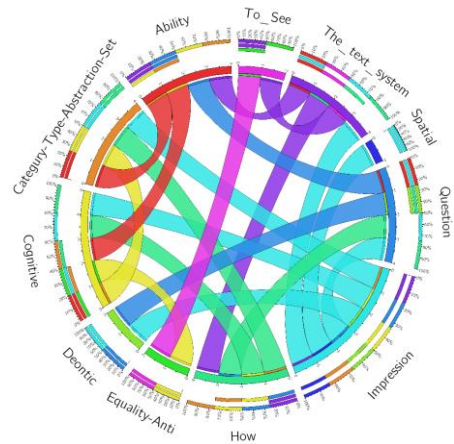

**b)**

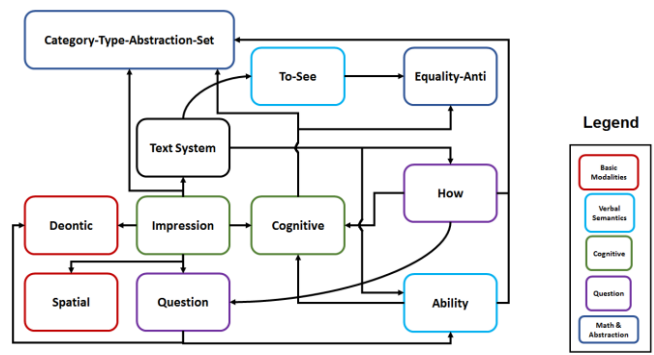

**Figure 9-** the theories and the dependencies between them, for SUS-2. Depicted in two different visualization formats: A) Chord Diagrammatic Layout (Krzywinski et al., 2009), B) Connectionist Layout.

---

[4] For the transcript of the mentioned TED talk see:
https://www.ted.com/talks/andrea_ghez_the_hunt_for_a_supermassive_black_hole/transcript?language=en

> **Semantic Logic-2:**
> **Number of related theories: 11**
> **Number of dependencies between theories: 24**
> **Number of Semantic Model Elements:  14**
> **Number of  Operators\Modalities:   8**
> **Number of semantic axioms\rules:  15**
>
> **The-Text-System-Theory:**
> #S = How(Ability(See(Unseen)))
> #S  is-in  Beginning
>
> **To-See-Theory**
> See  <>  Unseen
>
> **Ability-Theory:**
> Ability  is-a Positive-Sense
>
> **How-Theory:**
> How  is-a  Positive-Sense
> How  is-a Question
>
> **Question-Theory:**
> Question(Ability(S))  =>  Abduction(P(Not(S)))
>
> **Cognitive-Theory:**
>  (A  <>  B)  And  (A(B))   =>    Wonder(A(B))
> A(B)  And  Wonder(B)  And  A  is-a  Positive-Sense    =>    Wonder(A(B))
>
> **Impression-Theory:**
> (S is-in  Beginning)  And  (Wonder(S))  => Attention-Policy
>                                                                          and    Excitement
> (Question(S) is-in  Beginning)  =>  Engagement
> Engagement   => Attention-Policy
> Excitement  =>  Attention-Policy
> Engagement  And  P(Not(Excitement)) =>  P(Not(Promotion))
>                                                                       And  P(Not(Excitement))
> Engagement   And  Excitement => Propagation

   Elicitation, extraction and definition of these rules and symbols take about 10 man-hours. It is important to notice that Semantic Logic and theories are reusable artifacts (like ontologies, meta-models and models) and in a real academic or commercial solution, a rule-base (or Semantic Logic base) could amortize and decrease the needed effort and overhead of Semantic Logic definition.

There is a special symbol for SUS in this Semantic Logic: #S. the SUS-2 is quantized by basic intuitions of the provided Semantic Logic; its specification in terms of basic intuitions is: How(Ability(See(Unseen))). This symbolic construction then was used as an input feed to start a chain of deductions and derivations with application of axioms of the provided Semantic Logic. The result of this process was a semantic lattice that its symbolic and abstract structure is provided in figure 10. The deduction process was automated by a program written in Java Language.

The result of automatic deductions is presented in figure 11. These complex values (structures of symbols) are result of our deduction process. Presence of Each of them in the result could be viewed as a measure of semantic load (and a semantic complexity element). For example, SUS-1 (in section 4) has no semantic load of "wonder" or "engagement". But SUS-2 have both of them. The configuration of values by a schema (like the one used for SUS-1) could be applied to the abstract lattice provided in figure 10. But for SUS-2, it is more attractive to consider the resulted semantic loads.

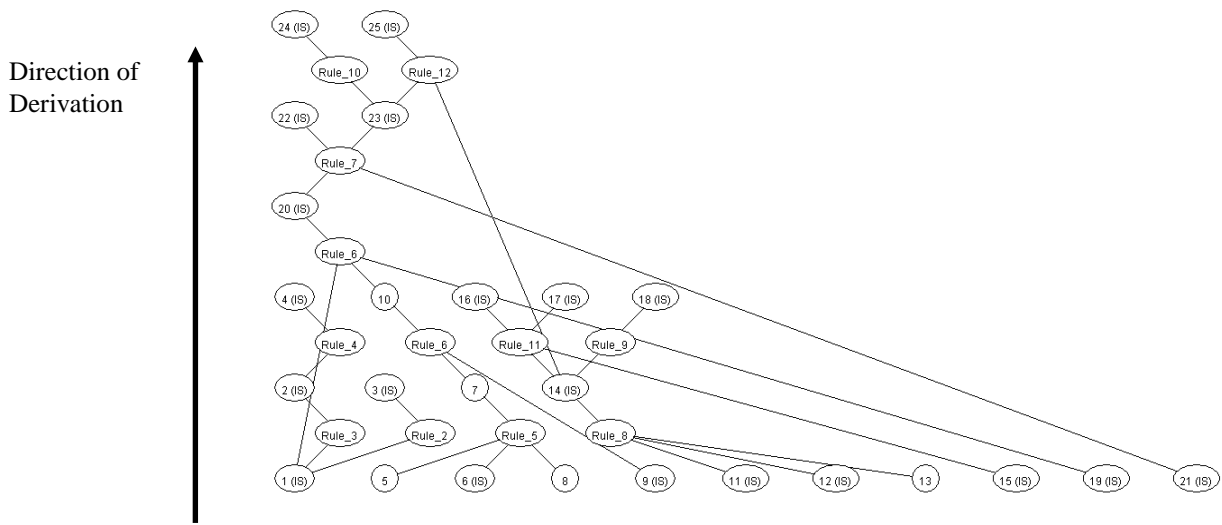

Direction of Derivation

**Figure 10-** derived and deducted abstract "semantic lattice" of SUS-2.

```
Deductions In System ==========================================
1    P ( Not ( Excitement ) )
2    Attention_Policy ( Excitement ( Wonder ( How ( Ability ( See ( Unseen ) ) ) ) ) )
3    Wonder ( How ( Ability ( See ( Unseen ) ) ) )
4    Attention_Policy
5    Positive_Sense ( Ability ( See ( Unseen ) ) )
6    Question ( Ability ( See ( Unseen ) ) )
7    Excitement ( Wonder ( How ( Ability ( See ( Unseen ) ) ) ) )
8    Engagement ( Ability ( See ( Unseen ) ) )
9    Abduction ( P ( Not ( See ( Unseen ) ) ) )
10   Attention_Policy ( Engagement ( Ability ( See ( Unseen ) ) ) )
11   P ( Not ( Promotion ) )
12   Propagation ( Engagement ( Ability ( See ( Unseen ) ) ) )
```

**Figure 11-** the result of automatic deductions by application of axioms of Semantic Logic for SUS2. Note: P is the deontic modality (operator) of possibility.

### 6.2- Experiments of Human Judge Evaluation

A series of test scenarios, based on human judgement experiments, are designed and proceeded to evaluate some aspects of the proposed method. An overview of these test scenarios and their characteristics are provided in Table-1.

The experiments of these test scenarios were run on a web-based platform: a mesh-up of web-based questionnaire services and mobile social network channels and groups. One paper-based questionnaire and 18 online (web-based) questionnaires are used to gather the human judgments data. The result is somehow a semantic-complexity, human-judge corpus which could be discussed in details and analyzed in a separate research paper.

**Table 1-** An Overview of Test Scenarios and Their Characteristics

|   | description | Number Of Participants | Number of Experiments | Overall Number of Human judgments | Evaluation Goal |
|---|---|---|---|---|---|
| **Test Scenario-1** | One Sentence, a lot of Participants | Totally 3198 | 3 | More than 12000 | Validity |
| **Test Scenario-2** | Some Sentences, One Domain, Many Participants | Totally 4354 | 16 | More than 15000 | Generality |

Based on an online questionnaire web-service platform, some hyperlinks are generated for accessing the online questionnaire forms. These links were propagated in some communities of mobile-social-networks (Communities of some subscribing channels of Telegram Messenger with 10K, 100K and 1M populations). Our public invitation content-posts have been reach totally Over 500K seen, 30K link-click and 7K different-participations. So a 2-node mesh of cloud-based services (questionnaire-platform service and the Messenger Service) shaped our instrumentation infrastructure of our Experimentation. It's a simple but proper example of a cloud-based solution for gathering 5V (or at least big) human-Judgment data.

**6.2.1. Test Scenario-1: Mutation Experiment**

Based on SUS-2, four meaningful mutant sentences are generated. Then three survey experiments have been run to ask people about comparison of semantic load and semantic complexity of these mutant sentences and SUS-2 (they didn't know which one is SUS-2). Because the main sentence and its mutants are from same semantic context, the comparison between them is more fare. In the first experiment, 22 computer students answered the questionnaire. In the second experiment, 92 arbitrary people participated in an online questionnaire from the web. In the third experiment, 3084 arbitrary people participated in our online questionnaire. The community of these three experiments were fully distinct and different from each other.

In these experiments, the participant persons have been asked to fill a comparison structure for judgment between the five sentences (see figure 12). In each comparison step, the more complex sentence should bubble up. Four comparison-steps (a, b, c and d) led each participant to his\her opinion about the answer: the most top sentence in the comparison structure have had the most semantic load and complexity.

Judgment experiments on the five sentences also have been done automatically by applying DAST method (the provided Semantic Logic in previous section applied for automatic deduction of semantic load and semantic complexity of these sentences). In a comparison step, if the DAST complexity value of two choices became equal, both of them were considered as the result of DAST judgment (Less than 10% of DAST judgments resulted in more than one sentence).

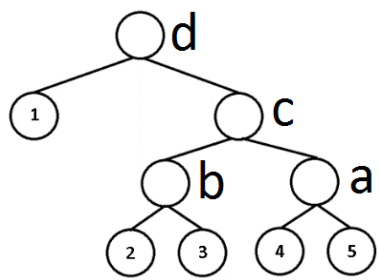

**Figure 12-** The comparison structure between the five sentences, in the questionnaire of the human judge experiment.

In the first experiment, all of students were agreed on all axioms of provided Semantic Logic. In the second Experiment, there was an agreement on 90% of axiom-agree-cases (each axiom-agree-case was constituted by a participant agreement with an axiom). This was 88% in the third experiment. So, the provided Semantic Logic was an axiomatization of their common sense (or common knowledge).

The questionnaire was not trivial: for example, the students were not fully agreed about answer of the questions. Majority of students in experiment-1 (64%) were agreed on SUS-2 (among 5 sentences) for its more semantic load and more semantic complexity than other sentences. In experiment-2, majority of participants (78%) were agreed on SUS-2 for its more semantic load and more semantic complexity than other sentences. This majority was 70% in the third experiment. These results are consistent with the results of DAST method automatic judgment: By applying the provided Semantic Logic, the more semantic load and complexity were deducted for SUS-2 than other sentences. A summary of results for two experiments are provided in Table-2. Please note that in our evaluation, Precision is defined as below:

**Definition 7.**

$$Precision = \frac{\text{TruePositive}}{\text{TruePositive} + \text{FalsePositive}}$$

Each comparison step (comparison between two sentences) is a judgment. In our evaluation setting, each match between DAST judgment (DJ) and Human Judgment (HJ) is considered as a TruePositive.

Considering (a, b, c, d) components for each 5-sentence comparison result (in accordance to figure-12), we could denote the operational definition (Kant & Srinivasan, 1992) of the precision concept as the below formulas:

**Definition 8.**

$$OverallResult\_Precision(DJ, HJ) = \frac{\sum_{i=0}^{|HJ|} EqualityValue(DJ.d, HJ_i.d)}{|HJ|}$$

$$NoDeviation\_Precision(DJ, HJ) = \frac{\sum_{i=0}^{|HJ|} VectorEqualityValue(DJ, HJ_i)}{|HJ|}$$

$$CompSteps\_Precision(DJ, HJ) = \frac{\sum_{i=0}^{|HJ|} \sum_{q=a}^{d} EqualityValue(DJ.q, HJ_i.q)}{4 \times |HJ|}$$

$$EqualityValue(x, y) = \begin{cases} 1 & , x = y \\ 0 & , x \neq y \end{cases}$$

$$VectorEqualityValue(x, y) = \begin{cases} 1 & , (x.a = y.a) \wedge (x.b = y.b) \wedge \\ & (x.c = y.c) \wedge (x.d = y.d) \\ 0 & , Otherwise \end{cases}$$

**Table 2-** A Summary of Experiments Results of Test Scenario-1

|  | Number of Participants | Participants Agreement on Axiom-Agree-Cases | Precision For Overall Result | Precision For No Deviation In Comp. Steps | Precision For Comp. Steps |
|---|---|---|---|---|---|
| **Experiment-1** | 22 | 100% | 64% | 45% | 81% |
| **Experiment-2** | 92 | 90% | 78% | 62% | 83% |
| **Experiment-3** | 3084 | 88% | 70% | 54% | 81% |
| **Total of Experiment-1 & 2** | 114 | 92% | 75% | 59% | 83% |
| **Total of All Three Experiments** | 3198 | 88% | 70% | 54% | 81% |

With an about 70% Precision for overall result (comparative with 20% for random baseline), these experiments reveal some levels of validity for DAST Method in mimicking and computing the human judgments of semantic complexity comparisons.

In figures 13 and 14, some resulting diagrams and curves from the experiments data are provided. Pseudo-normal, power-law-like and exponential curves are among them.

The distribution of percentage share for choices of each experiment is a pseudo-normal curve (For experiment 1, 2 and 3, these curves are depicted in figure 14.c, 14.f and 14.i.).

In each comparison step (figure-12), there are two choices. So there are 16 combinatorial paths in the space of comparisons. For the human judgment data from the three experiments, the percentage share of each of these 16 paths (in the ascending order) are depicted in the figures 14.b, 14.e and 14.h. the resulting curves are suggesting a power-law-like distribution.

The Analyze of Answers to each step of comparison is also noticeable. In experiment-3, the diagram of Percentage Share of Opinion Portions versus Number of Deviated Steps shows a meaningful exponential correlation (with $R^2=0.97$, see figure 13).

In experiment-1, about 45 percent of students didn't have any deviation in comparison-steps regards the decisions of DAST method. About 42 percent had deviation only in one step. Other deviated comparison paths had about 5, 9 and 1 percent share of opinion portions[5]. These values shape up an exponential descending curve (see figure 14.a).

In experiment-2, about 62 percent of participants didn't have any deviation in comparison-steps regards the decisions of DAST method. About 23 percent had deviation only in one step. Other deviated comparison paths and results had a 7, 7 and 2 percent share of opinion portions. These values shape up an exponential descending curve (see figure 14.d).

In experiment-3, about 54 percent of participants didn't have any deviation in comparison-steps regards the decisions of DAST method. About 33 percent had deviation only in one step. Other deviated comparison paths had about 7, 5 and 2 percent share of opinion portions. These values shape up an exponential descending curve (see figure 14.g).

In order to interpret and explain the reason of these exponential descending curves, a Markovian Model for the process of common-sense multi-steps semantic-complexity reasoning in people is introduced (see figure 15). In each step, the probability of reasoning consistently with common-sense (the $\beta_i$ values) is greater than the probability of a deviation or noise regards common-sense (the $\alpha_i$ Values). So the mode of the distribution (or may be the majority _like our case_, depending on $\alpha_i$ and $\beta_i$ values) is consistent with deductions of (and based on) common-sense knowledge.

Participants were asked to tell us whether they are agree or disagree with each of axioms of provided Semantic Logic. This is named as their consensus on the provided semantic logic. Based on it, the participants were partitioned in 6 separate classes: people with 0%, 20%, 40%, 60%, 80% and 100% agreement on the provided semantics axioms. These classes designate 6 different level of consensus on the provided common semantics. Based on comparison of DJ and HJ data, the members of each different class had their "percentage of agreement with DAST overall result". The relation between "consensus on semantics axioms" and "percentage of agreement with DAST overall result" is somehow interesting (see figure 16). It suggests that the consensus on semantics-axioms has a sigmoid-like triggering effect on the semantic-complexity-agreement.

It suggests a Gate-Mechanism (Like the role of activation functions in artificial neurons) for semantic-complexity-control in an extended cognitive system in behalf of a community of people. We assume and suggest that a network of these Gate-Mechanisms could shape a cybernetics technique for constructing "cognitive expert systems for semantic complexity purposes".

---

[5] Please note that the fractional part of percentages are rounded.

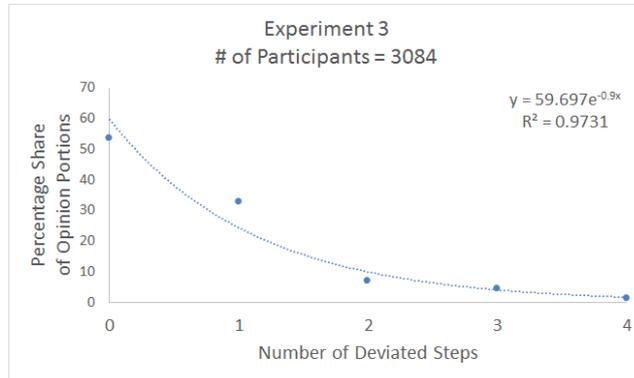

**Figure 13-** Exponential Curve Regression for Percentage Share of Opinion Portions versus Number of Deviated Steps.

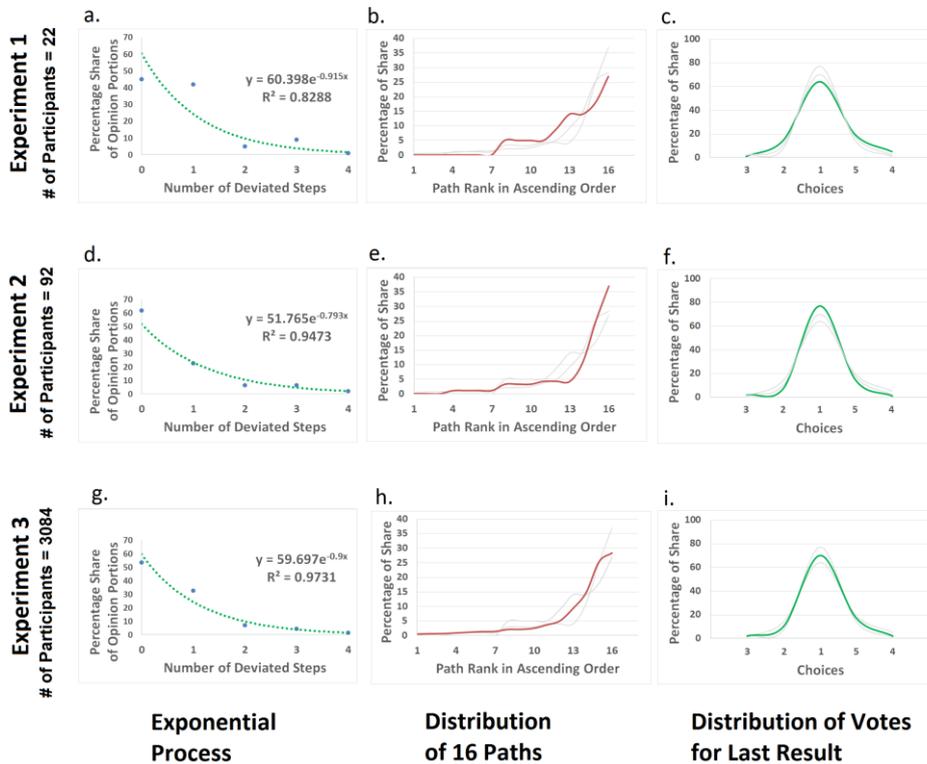

**Figure 14-** exponential curves for share of opinion portions related to Number of deviated steps (a, d and g). Distribution of percentage of share for 16 paths (b, e and h). Distribution of percentage share for choices of each experiment (c, f and i). Experiment-1, with 22 Participants. Experiment-2, with 92 Participants. Experiment-3, with 3084 Participants.

$$\forall i \, . \, \alpha_i < \beta_i$$

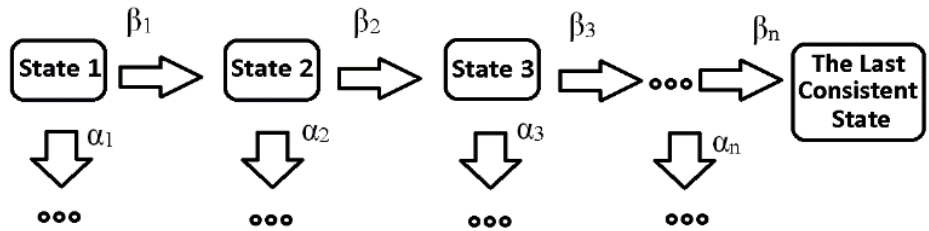

**Figure15** - A Markovian model for the process of common-sense multi-steps semantic-complexity reasoning in people

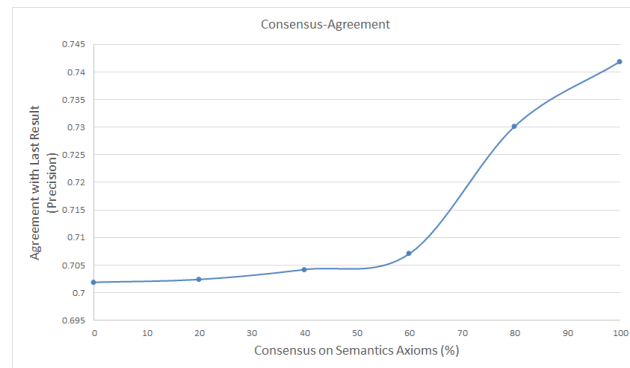

**Figure 16-** semantic-complexity-agreement versus consensus on semantics-axioms.

### 6.2.2. Test Scenario-2: Semantic Complexity Comparisons

Based on the previous comparison structure (depicted in figure 12), a set of online-questionnaires were designed and 16 different experiments were run in this test scenario. Each experiment involves its 5 sentences from classic Persian poems (so a total of 80 sentences in all 16 experiments). Sentences in each 5-sentences-tuple have similar key concepts but different meanings (hence from a common meaning domain or semantic filed). Participants were asked to compare sentences of each set in the same manner (in the step-by-step comparison structure of figure 12). A simple depiction of a sample questionnaire is provided in figure 17.

**Figure 17-** A simple depiction of a sample questionnaire and sample of its filling.

More details of these experiments and questionnaires are provided in DAST Dataset (Besharati & Izadi, 2019).

Judgment experiments on the 5-sentences sets also have been done by applying DAST method. The human judgments data was compared with the DAST judgments. A summary of results for these 16 experiments are provided in Table-3.

These Experiments-result demonstrates that our method consistently outperforms the baseline (random selection between five sentences). Precision for Overall Result has a Mean 58% (comparative with 20% for random baseline). Precision for No Deviation in Comparison Steps has a Mean 31% (comparative with 13% for random baseline). With these results, we see repeated validities. These experiments reveal some levels of generality for DAST Method in valid evaluation of semantic complexity. So DAST Method has an examination with repeated validity and not just a singularity of a very special case.

In each experiment, participants shaped a vote list for 5 sentences. Each sentence in each experiment, had a vote-value: the portion of participants that selected that sentence as the most complex one in the regarding 5-sentence-tuple. Overall, 80 sentences in 16 experiments received 80 vote-values.

DAST method, also, computes a vote-like value: semantic complexity relative-values. By semantic complexity relative-value we mean the semantic complexity value of a sentence divided by the total of semantic complexity values of all 5 sentences in its 5-sentence-tuple.

As a hypothesis, vote-values may be correlated with DAST computed semantic complexity relative-values of these sentences.

**Hypothesis 1.** Vote-values are correlated with DAST semantic complexity relative-values.

The correlation with vote-values (as an experimental indication of semantic complexity index of sentences) introduces the DAST semantic complexity as an indication of semantic complexity of sentences. For an estimation of semantic complexity, we could consider Vote-values and its correlated metrics (such as DAST semantic complexity).

Table 3- A Summary of Experiments Results of Test Scenario-2

|  | Number of Participants | Precision For Overall Result | Precision For No Deviation In Comp. Steps | Precision For Comp. Steps |
|---|---|---|---|---|
| **Experiment-1** | 172 | 59% | 26% | 61% |
| **Experiment-2** | 675 | 65% | 20% | 70% |
| **Experiment-3** | 154 | 46% | 21% | 64% |
| **Experiment-4** | 196 | 48% | 41% | 67% |
| **Experiment-5** | 160 | 46% | 22% | 63% |
| **Experiment-6** | 748 | 77% | 54% | 70% |
| **Experiment-7** | 144 | 62% | 26% | 57% |
| **Experiment-8** | 416 | 64% | 46% | 73% |
| **Experiment-9** | 177 | 70% | 41% | 71% |
| **Experiment-10** | 148 | 48% | 40% | 57% |
| **Experiment-11** | 387 | 73% | 20% | 55% |
| **Experiment-12** | 140 | 45% | 31% | 59% |
| **Experiment-13** | 203 | 81% | 36% | 76% |
| **Experiment-14** | 328 | 83% | 58% | 76% |
| **Experiment-15** | 99 | 41% | 34% | 59% |
| **Experiment-16** | 207 | 68% | 17% | 49% |
|  | Total=4354 Avg.=272 | Avg.=61% STDEV=13 | Avg.=33% STDEV=12 | Avg.=64% STDEV=8 |

In figure 18, The 80 vote-values for 80 sentences of these 16 experiments are depicted versus DAST semantic complexity relative-values. The regression shows a meaningful linear correlation between vote-values and DAST semantic complexity relative-values ($R^2$=0.83). So the general claim of Hypothesis 1 has been supported by the results of this experiment.

In figure 19, Precisions for overall result are depicted versus Precisions for Comparison Steps. This shapes up 16 data points (one for each experiment). Two data points are excluded (regarded as outlier) and the trend of the other fourteen data points are depicted. The result of cluster regression suggests that an extrapolated extremum value on (0.79, 0.83). Roughly, 80% precision in micro-steps results in an extremum 80% precision in macro-state. It is a pseudo-factually result.

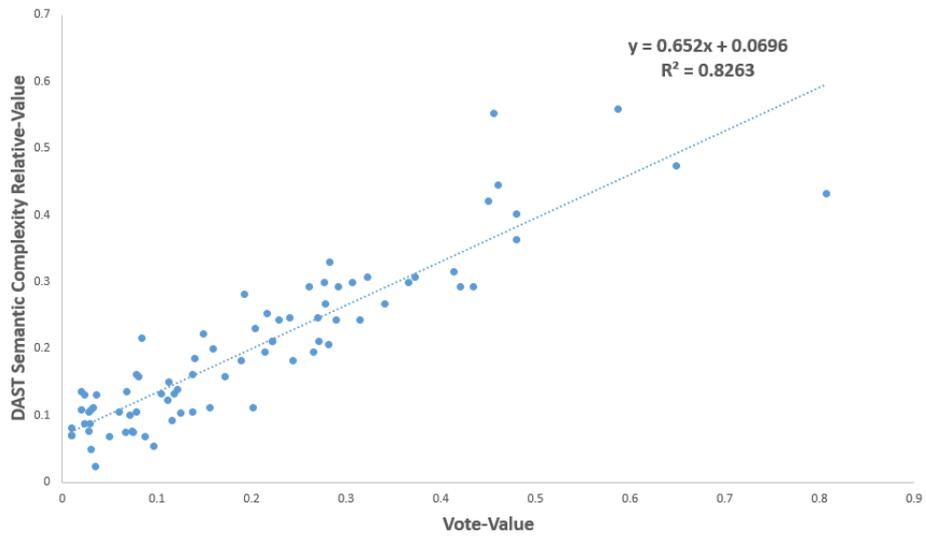

**Figure 18-** The linear correlation between Vote-values and DAST semantic complexity relative-values, depicted for 80 data-points regarding 80 sentences of experiments.

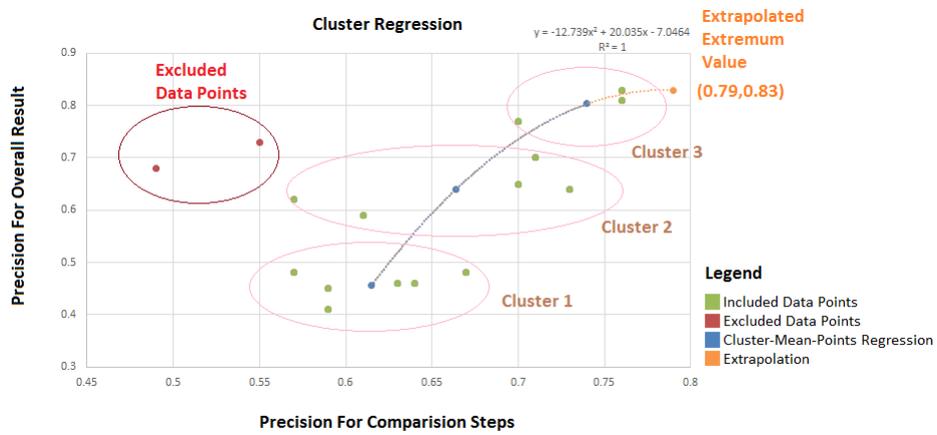

**Figure 19-** Precision for overall result versus Precision for Comparison Steps. The result of cluster regression suggests that an extrapolated extremum value on (0.79, 0.83).

### 6.3- Corpus Based Evaluation

One of main applications of proposed method is in the field of Automatic Readability Assessment (ARA). In order to compare our proposed method with the state of the art in this field, we reuse the data available from a corpus from a previous research: Scanpath Complexity Dataset (Mishra, Kanojia, Nagar, Dey, & Bhattacharyya, 2017). This dataset is containing sentences with level of readability labels and values. As a gold standard or benchmark, this data could be used to assess the performance of our method in separating easy and hard sentences.

This dataset involves calculated values of numerous readability formulas. So after all, we could compare our method (as a procedure for semantic readability assessment) with other readability formulas. This is an important evaluation because others have a syntax-biased approach and our approach is a semantic one. So the opportunities of semantic approaches for improving the state of the art may be revealed.

### 6.3.1. Evaluation based on Scanpath Complexity Dataset

This dataset contains 32 paragraphs of 50 − 200 words on 16 different topics belonging to the domains of history, geography, science and literature (Mishra et al., 2017). For each topic, there are two comparable paragraphs from Wikipedia and Simple Wikipedia.

This dataset contains two sets of attributes for each paragraph: 1) calculated values of linguistics measures and readability formulas, and 2) Eye fixation durations (gaze data) and readability annotations scores calculated from reading experiments by 16 participated human subjects (From the eye-tracking experiments, this dataset involves 512 unique eye scanpaths from 16 participants, each reading 32 paragraphs. Also, the paragraphs are annotated by these 16 participants in 5 readability levels).

In order to evaluate DAST Method, we consider the concept of "# of involving semantic theories" as a first level estimation of semantic complexity of a text (see Definition 6) and we measured this for all paragraphs of Scanpath dataset. Then these DAST-measured semantic-complexity estimations compared with complexity scores provided in the given dataset, including: 1) eye fixation durations (as a measure of semantic complexity), 2) readability formula scores and 3) annotated scores for readability level by human subjects.

We need to define a key concept in order to prepare our method of evaluation and comparison. For each topic, this dataset contains two comparable paragraphs from Wikipedia and simple Wikipedia (Mishra et al., 2017). So we have 16 simple-hard paragraph-pairs. For each paragraph-pair, based on a text complexity evaluation function F, we define Difficulty Ratio as this:

**Definition 9.**

$$\text{F-Difficulty-Ratio}_i (\text{Simple}_i, \text{Hard}_i) = \frac{\text{F}(\text{Simple}_i)}{\text{F}(\text{Hard}_i)}$$

This ratio measures the distinction power of function F on the (Simple$_i$, Hard$_i$) paragraph-pair. For each evaluation function (or readability formula), we could compute 16 data points regarding to 16 paragraph-pairs of the Scanpath dataset.

There are two set of evaluation function (or Readability Formula) values in our analysis in this section. The first set of function values which are calculated based on our experimentation: DASTEX and DAST Evaluation Time (the amount of time that was needed for a human expert to run DAST evaluation process on the paragraph). The second set of function values are obtained from the original research of the dataset (Mishra et al., 2017): Fixation Time, Word Count, Readability Level, Gunning-Fog Score, Flesch-Kincaid Grade Level, SMOG Index, Automated Readability Index, Coleman-Liau Index and Average Grade Level (the average of the five previously mentioned readability scores in this list). So our contributed data parameters are DASTEX and DAST Evaluation Time which are calculated for all 32 paragraphs of the dataset. After our experimentation, we calculate DR-values for all of these 9 previous and 2 new evaluation functions of the dataset.

### 6.3.1.1. Evaluation based on eye fixation durations

DAST Method proposes its Semantic Index (Definition 6) as an estimation of semantic complexity. So it should have some distinction power in separating hard and simple paragraphs in each paragraph-pair. We could consider DR-values (Definition 9) for the sake of this evaluation goal.

More Complex a text, More Time Need to Read it. So in order of an estimation of semantic complexity, we could consider eye fixation duration time (Measure1 in Definition 10). Alternatively, it may be proper to consider its intensity (Measure2 in Definition 11).

The more two paragraphs are different in their semantic Complexity, the more they are different in Fixation duration times, and the less calculated value for Fixation durations-DR. Alternatively, The more two paragraphs are different in their semantic Complexity, the more they are different in DAST Semantic Complexity Index (DASTEX), and the less calculated value for DASTEX-DR.

Put all things together, DASTEX-DR and eye Fixation durations-DR for paragraph-pairs have a correlation, this is our experiment hypothesis in this evaluation step.

| | | | |
|---|---|---|---|
| **Definition 10.** | Measure1: | Fixation Duration Time | |
| **Definition 11.** | Measure2: | Fixation Duration Time per Word | |
| **Hypothesis 2.** | Fixation durations-DR | ≈ | DASTEX-DR |
| **Hypothesis 2.a** | Measure1-DR | ≈ | DASTEX-DR |
| **Hypothesis 2.b** | Measure2-DR | ≈ | DASTEX-DR |

We calculate the DASTEX-DR values for every of 16 paragraph-pairs by running our experiment. For this, a human expert analyzed the paragraphs to enumerate involving semantic theories. Based on DAST model, a theory must have its basic intuitions in the text (Concepts, Entities, Objects, Realities, Identities, ... in the term of their words or senses) and also some symbolic rules could be found (or defined) to elaborate underlying knowledge about them. As an example, some involving semantic theories (and some samples of their intuitions and rules) of a paragraph are provided here (see Table-4).

A linear regression on DASTEX-DR versos Measure1-DR and Measure2-DR was done to test the Hypothesis 2. When all of 16 data points were considered, the precision of linear regression was not acceptable: not Measure1-DR nor Measure2-DR didn't support an accurate linear relation with DASTEX-DR. it was a negation result for Hypothesis 2. But further analysis reveals there was also something else. By splitting the 16 data points into two 8-elements distinct classes (based on the genre of the belonging paragraph topic), it found that for "Class1: history and literature topics", the Measure1-DR have an accurate linear relation with DASTEX-DR ($R^2$=0.98). Furthermore, an accurate linear relation (by excluding one exceptional data point which is regarded as outlier) was found between Measure2-DR and DASTEX-DR for "Class2: geography and science topics" ($R^2$=0.96). See figure 20 for further details.

After all, the experiment results support Hypothesis 2.a for "Class1: history and literature topics" and Hypothesis 2.b for "Class2: geography and science topics". So, based on a genre-aware viewpoint, general claim of Hypothesis 2 has been supported by the results of this experiment. This also proposes a new opportunity for designing a new "soft sensor" for genre detection, based on semantic complexity indications of DASTEX.

**Table 4-** A paragraph from dataset and some involving semantic theories.

| **Paragraph Under Study:** ||| 
|---|---|---|
| A Hilbert space is a mathematical concept which is a more general type of Euclidean space. It takes the mathematics used in two and three dimensions, and asks what happens if there are more than three dimensions. It is named after David Hilbert. |||
| **Semantic Theory ID** | **Intuitions Words** | **Some Rules on Principal Intuitions** |
| 1 | Hilbert space | Hilbert [(is-a)] math-scientist <br> Hilbert space [(is-owned-to)] Hilbert |
| 2 | Space | Space => Many(GeometricPoint) |
| 3 | more | (A => B) And (more(A)) => P(more(B)) |
| 4 | Mathematical, Euclidean space | Mathematical => Math <br> Math => Number \| Value \| Ratio \| Count \| Abstract ID <br> Euclidean space => Some(Dimension) <br> Dimension => Value |
| 5 | Concept, general, type | Concept [(is-a)] Cognitive-Term <br> General => Abstraction <br> Type => Abstraction <br> Abstraction [(is-a)] Cognitive-Term |
| 6 | Ask | Ask [(is-a)] Verb <br> A Ask B => (B [(is-a)] Question) And (A Need Answer(B)) |
| ... | ... | ... |

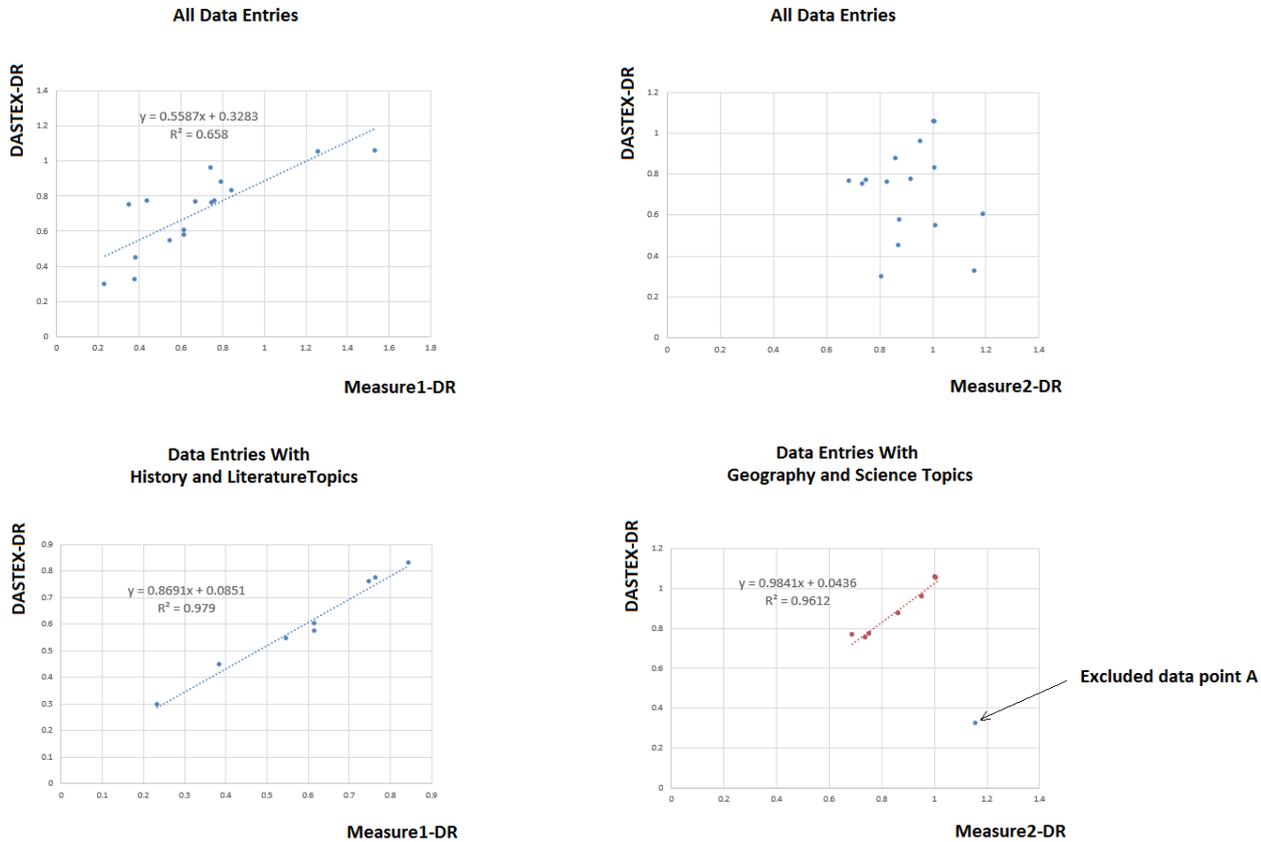

**Figure 20-** Linear Regression Results for DASTEX-DR versos Measure1-DR and Measure2-DR.

### 6.3.1.2. Evaluation based on readability formula scores

Considering the complexity evaluation function F (or Readability Formula F), The distribution of F-DR values on all paragraph-pairs in Ascending order (which we call it as DR Curve) could be considered as a fingerprint of F. Being based on a certain dataset, different readability functions have different DR-fingerprints. We could construct and compare these fingerprints (i.e. DR Curves) in order to compare the overall distinction behavior of different readability functions. The result of such an analysis is provided in figure 21. In this figure, All DR-values are uniformly scaled to a maximum of 1.

Three clusters of curves are found in this analysis (see figure 22 for independent depiction of these clusters). So based on the scaled value of DR-Curves, there are three classes of overall distinction behavior among readability formulas and functions. DASTEX-DR is in a somehow good cluster: because its cluster-co-members are DR-values of two important readability measures: Word Count (i.e. Text Length) and Fixation Time (i.e. Read Time).

Word count is an objective measure and Fixation Time is a subjective one. So DASTEX-DR has a hybrid simulation power for both some objective and subjective readability measurements. Based on DR-Curves Clusters, other readability formulas are not co-member with important word-count and fixation-time measures, but DASTEX is. This is a special characteristic for DASTEX among other readability formulas, albeit based on DR-Value analysis.

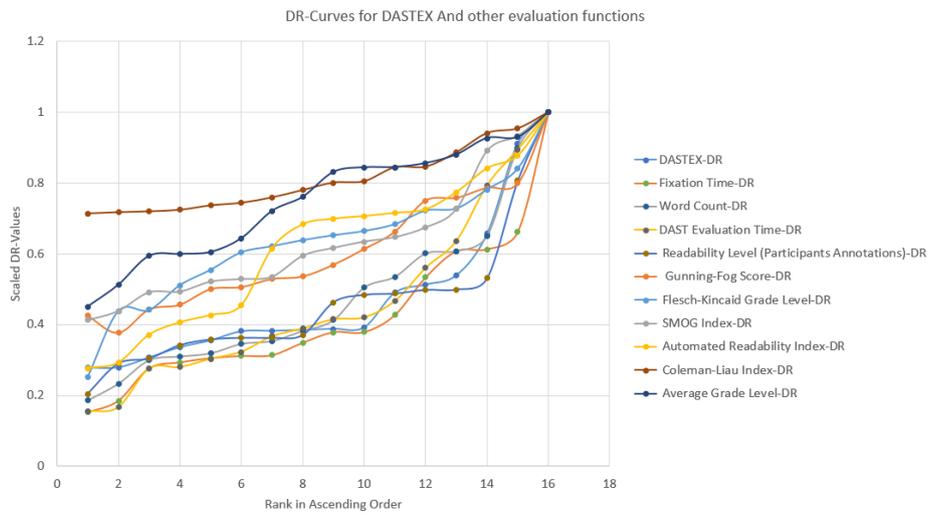

**Figure 21-** DR-Curves for DASTEX and other evaluation functions. Three clusters of curves are found.

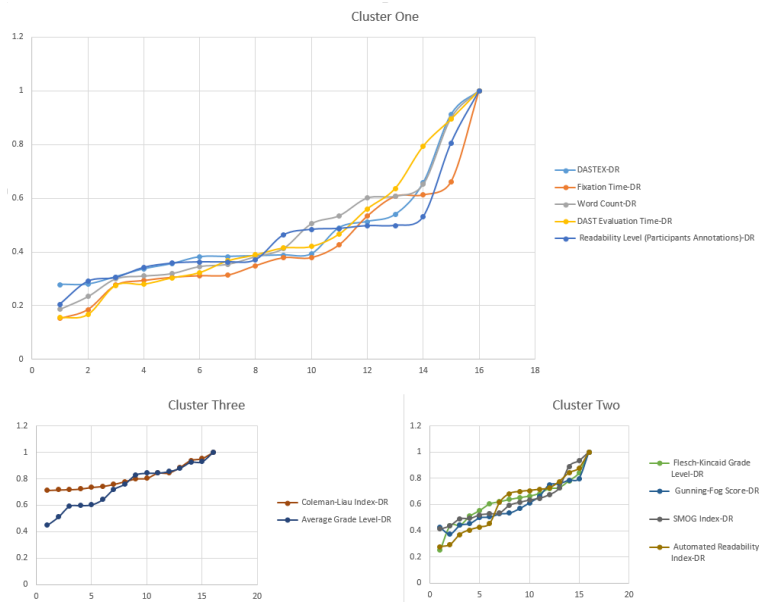

**Figure 22-** Independent Depiction of the Three DR-Curves Clusters.

### 6.3.1.3. Evaluation based on Overall DR of Functions

Each evaluation function has an overall distinction power due to its average values for simple and hard paragraphs. For an evaluation function F:

$$\text{F-Overall Difficulty Ratio} = \frac{\text{(Average of F on Simple Instances)}}{\text{(Average of F on Hard Instances)}}$$

In table-5, averages of DASTEX values and 8 other measures (for 16 paragraphs from Wikipedia and 16 paragraphs from simpleWikipedia) are provided. Calculated difficulty-ratios for these evaluation functions show a proximity of DASTEX distinction power (in term of Difficulty-Ratio) to Fixation-Time, Word-Count and Readability-Level (Participants Annotations).

DASTEX has a minimum error in its overall Difficulty Ratio due to Fixation-Time. Also, DASTEX has a second position in the lowest errors due to Word-Count and Readability-Level (See figure 23 for a sorted depiction of error values).

**Table-5**. Parameter averages, Their Overall Difficulty Ratios and Their Error Percentages

| | DASTEX | Fixation Time | Word Count | Readability Level | Flesch-Kincaid Grade Level | Gunning-Fog Score | Coleman-Liau Index | SMOG Index | Automated Readability Index |
|---|---|---|---|---|---|---|---|---|---|
| Average for SimpleWikipedia Paragraphs | 24.19 | 31308.59 | 93.56 | 2.57 | 7.46 | 8.25 | 11.30 | 9.02 | 6.78 |
| Average for Wikipedia Paragraphs | 37.81 | 50367.07 | 132.19 | 4.39 | 14.26 | 15.22 | 14.31 | 13.36 | 14.79 |
| Difficulty-Ratio | 0.64 | 0.62 | 0.71 | 0.58 | 0.52 | 0.54 | 0.79 | 0.68 | 0.46 |
| Error Percentage (Based on Comparison with Average-Fixation-Time-DR) | 3.23 | 0.00 | 14.52 | 6.45 | 16.13 | 12.90 | 27.42 | 9.68 | 25.81 |
| Error Percentage (Based on Comparison with Average-Word-Count-DR) | 9.86 | 12.68 | 0.00 | 18.31 | 26.76 | 23.94 | 11.27 | 4.23 | 35.21 |
| Error Percentage (Based on Comparison with Readability-Level-DR) | 10.34 | 6.90 | 22.41 | 0.00 | 10.34 | 6.90 | 36.21 | 17.24 | 20.69 |

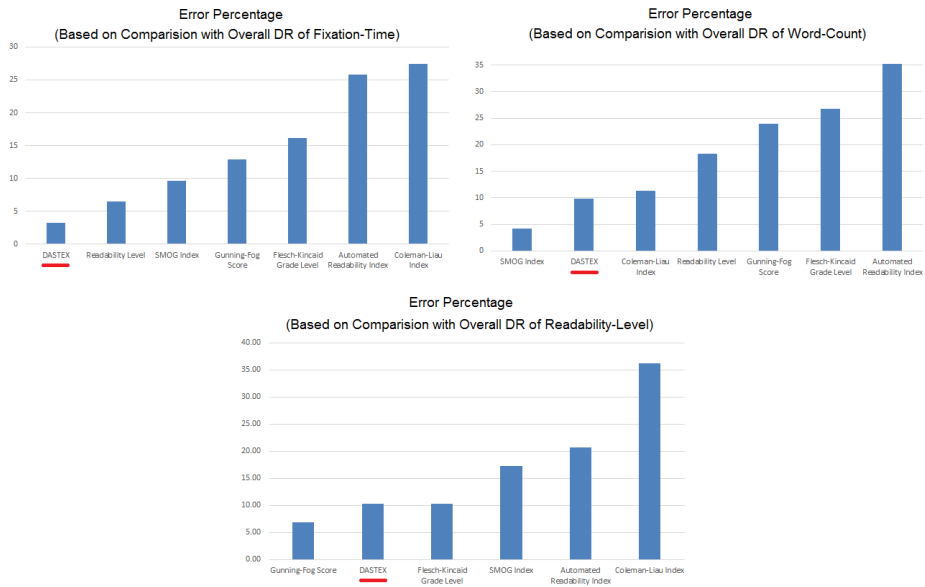

**Figure 23-** Error Percentage for different evaluation functions due to proximity of their overall distinction power to Fixation-Time, Word-Count and Readability-Level distinction powers (The DASTEX position has a red marker).

## 7. Discussion and Conclusion

There were some similar approaches in different research domains which incorporate some Atoms for knowledge structuring and meaning construction (atoms in Lisp for computational values, Objects in OO Languages in behalf of real-world identities, Ontos in Ontology approaches as a representation of real-world entities, terms and formula in logic-based approaches for truth construction, duality of state-and-transition-pairs in dynamic system modeling and etc.). We see all of these examples as different formulations of basic intuitions. However, "the creation of knowledge involves the coalescing of otherwise undifferentiated stimuli into representable forms" (Olsher, 2014) and for semantic modeling of a text, we need some sort of knowledge representation. By Comparing with our approach, Stimuli are intuitions and representation forms are symbols. Concisely, Representation of basic intuitions by symbols and representation of complex meanings by a construction of symbols (= a lattice) are basic principles of our method.

In the Cognitive Assistant research domain, Intuition is considered as a human-specific attribute of thinking (Fulbright, 2016). Human thinking and machine thinking with their different attributes and natures could collaborate and contribute to produce better results in cognitive assistant systems (Kelly III & Hamm, 2013) (Fulbright, 2016).

## 8. Future Works

Calculated Semantic Complexity on a Semantic Lattice could be interpreted as Fuzzy probability or possibility of that lattice and its associated concept or entity. Especially, in the case of abductive reasoning in cognitive or intelligence agents, Or Artificial Brains, this calculated semantic value could be valuable and useful. Also, a Fuzzy or Many-Valued approach to content-semantic could be supported by this interpretation of semantic Complexity Values.

A semantic entropy or complexity measure could be served in order to develop or evolve or construct the complex metaphors. Also, Kolmogorov Complexity of Metaphor could be calculated based on this underlying Logic. This Kolmogorov Complexity could be a Content measure or metaphor metric, both in literature studies and computational linguistics. By Machine-Learning of the occurring or lattice-Patterns of Metaphor Semantic Construction, a machine could be able to mimic the process of complex content or metaphor composition (in a compositional construction manner (Izadi, Movaghar, & Arbab, 2007) (Clarke, Proença, Lazovik, & Arbab, 2011) for content-generation processes). Albeit, the more quality in results with lower complexity in process could be achieved by human-supervision and human-intervention in the process of content generation. Each content system could be modeled or redefined in terms of its underlying logic (proof-view and construction process perspective to systems, ontos and objects under study). For example a successful film, story, poem, situation, system, event or any other intuitionistic reality could be redefined (or quantized) in terms of its underlying logic. Then, a learner system could mimic this success and yet evolves its trend or version. Or any other successful process (for example, in human-based systems or in social systems or in scientific research success stories or in industry success stories and cases) this method of logical interpretation and reprogramming could be established and give benefits (For example, an unformal book on a success story could be interpreted).

At last, we could conclude that the meanings of a given text are somehow computable. The DAST model could be served as a platform to facilitate such a meaning computation for Deciding About Semantic complexity of Texts.